\documentclass{article}

\PassOptionsToPackage{numbers, compress}{natbib}
\usepackage[preprint]{neurips_2024}

\usepackage{enumerate}
\usepackage[utf8]{inputenc} % allow utf-8 input
\usepackage[T1]{fontenc}    % use 8-bit T1 fonts
\usepackage{url}            % simple URL typesetting
\usepackage{booktabs}       % professional-quality tables
\usepackage{amsfonts}       % blackboard math symbols
\usepackage{nicefrac}       % compact symbols for 1/2, etc.
\usepackage{microtype}      % microtypography
\usepackage{xcolor}         % colors
\usepackage{amsmath}        % professonal math sign and equation
\usepackage{subcaption}     % subcaption

\definecolor{citecolor}{HTML}{0071bc}
\usepackage[colorlinks, linkcolor=red, colorlinks, anchorcolor=blue, citecolor=citecolor, pagebackref=True]{hyperref}

\usepackage{cleveref}       % clever ref
\usepackage{graphicx}       % graphicx
\usepackage{multirow}       % table

\definecolor{MyDarkBlue}{rgb}{0,0.08,1}
\definecolor{MyDarkGreen}{rgb}{0.02,0.6,0.02}
\definecolor{MyDarkRed}{rgb}{0.8,0.02,0.02}
\definecolor{MyDarkOrange}{rgb}{0.40,0.2,0.02}
\definecolor{MyPurple}{RGB}{111,0,255}
\definecolor{MyRed}{rgb}{1.0,0.0,0.0}
\definecolor{MyGold}{rgb}{0.75,0.6,0.12}
\definecolor{MyDarkgray}{rgb}{0.66, 0.66, 0.66}

\usepackage{array}

\newcolumntype{x}[1]{>{\centering\arraybackslash}p{#1}}
\newcolumntype{y}[1]{>{\raggedright\arraybackslash}p{#1}}
\newcolumntype{z}[1]{>{\raggedleft\arraybackslash}p{#1}}
\newcommand{\tablestyle}[2]{\setlength{\tabcolsep}{#1}\renewcommand{\arraystretch}{#2}\centering\footnotesize}

\definecolor{gray}{HTML}{efefef}
\title{Tailor3D: Customized 3D Assets Editing and Generation with Dual-Side Images}

\author{%
  Zhangyang Qi\textsuperscript{$1,2*$} \quad 
  Yunhan Yang\textsuperscript{$1*$} \quad
  Mengchen Zhang\textsuperscript{$2$} \quad
  Long Xing\textsuperscript{$2$} \quad
  Xiaoyang Wu\textsuperscript{$1$} \quad 
  \vspace{0.05cm} \\
  \textbf{
  Tong Wu\textsuperscript{$2,3$} \quad
  Dahua Lin\textsuperscript{$2,3$} \quad
  Xihui Liu\textsuperscript{$1$} \quad
  Jiaqi Wang\textsuperscript{$2\dag$} \quad
  Hengshuang Zhao\textsuperscript{$1\dag$} \quad
  }
  \vspace{0.1cm} \\
  \textsuperscript{$1$}The University of Hong Kong \quad
  \textsuperscript{$2$}Shanghai AI Lab \quad
  \textsuperscript{$3$}The Chinese University of Hong Kong
  \vspace{0.1cm} \\
  {\tt\small \url{https://tailor3d-2024.github.io/}} \quad {\small * equal contribution} \quad {\small $\dag$ corresponding author}
  \vspace{-0.2cm}
}

\begin{document}

\maketitle

\begin{figure}[h]
  \vspace{-7mm}
  \centering
  \includegraphics[width=0.98\textwidth]{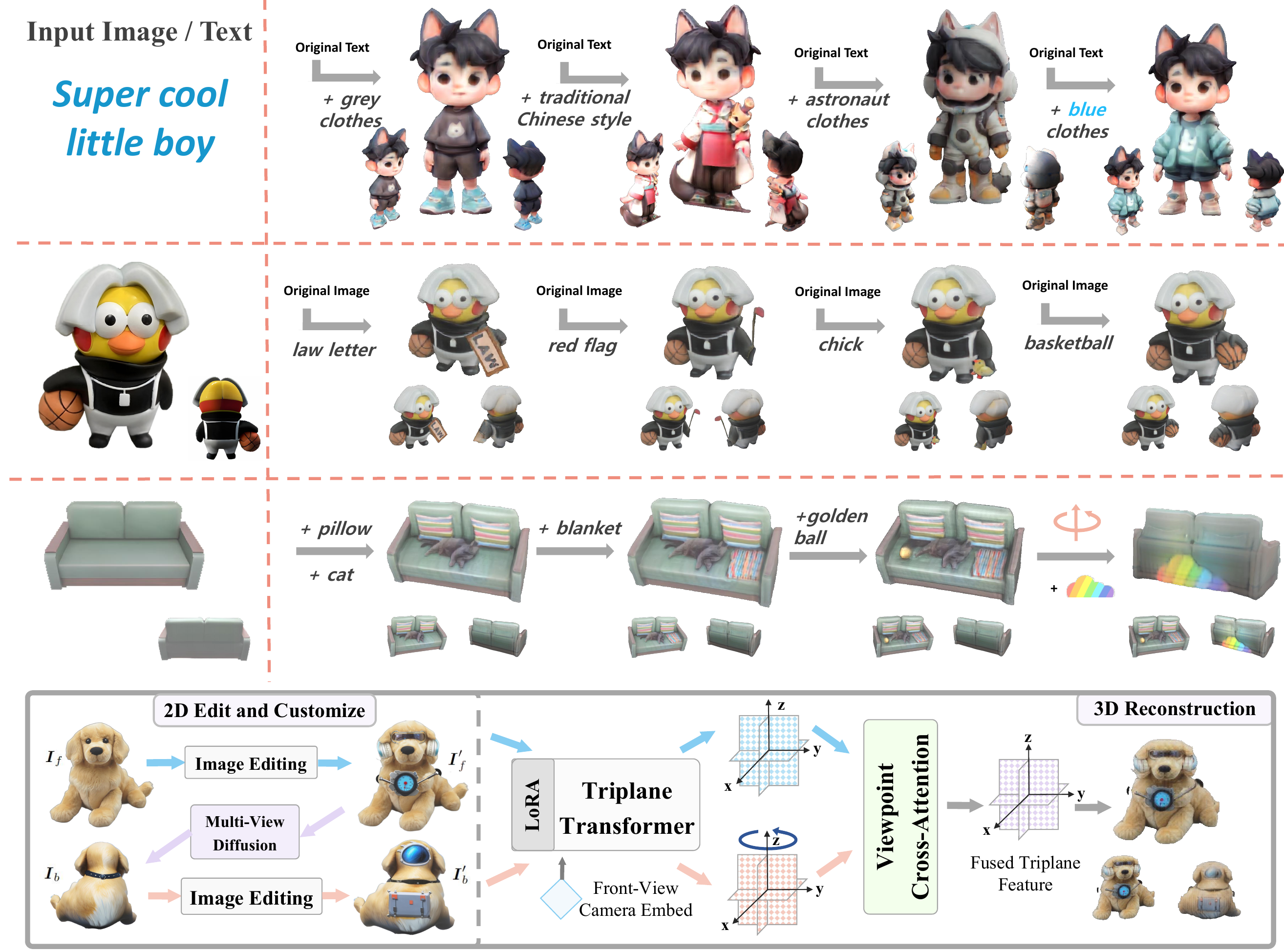}
  \caption{\textbf{Results and Pipeline}. We show our method for 3D style customization, as well as geometry and texture editing. Our pipeline involves editing images and generating the 3D object using Dual-sided LRM, with each step completed in just 5 seconds, allowing for rapid 3D object customization.}
  \label{fig_teaser}
\end{figure}
\vspace{-2mm}

\begin{abstract}
% Recently
Recent advances in 3D AIGC have shown promise in directly creating 3D objects from text and images, offering significant cost savings in animation and product design. However, detailed edit and customization of 3D assets remains a long-standing challenge. Specifically, 3D Generation methods lack the ability to follow finely detailed instructions as precisely as their 2D image creation counterparts. Imagine you can get a toy through 3D AIGC but with undesired accessories and dressing. To tackle this challenge, we propose a novel pipeline called Tailor3D, which swiftly creates customized 3D assets from editable dual-side images. We aim to emulate a tailor's ability to locally change objects or perform overall style transfer. Unlike creating 3D assets from multiple views, using dual-side images eliminates conflicts on overlapping areas that occur when editing individual views. Specifically, it begins by editing the front view, then generates the back view of the object through multi-view diffusion. Afterward, it proceeds to edit the back views. Finally, a Dual-sided LRM is proposed to seamlessly stitch together the front and back 3D features, akin to a tailor sewing together the front and back of a garment. The Dual-sided LRM rectifies imperfect consistencies between the front and back views, enhancing editing capabilities and reducing memory burdens while seamlessly integrating them into a unified 3D representation with the LoRA Triplane Transformer. Experimental results demonstrate Tailor3D's effectiveness across various 3D generation and editing tasks, including 3D generative fill and style transfer. It provides a user-friendly, efficient solution for editing 3D assets, with each editing step taking only seconds to complete.
\end{abstract}

\section{Introduction}
\label{sec:1_Introduction}
%% Intro: 2d aigc -> 3d aigc(optimized based) -> feed forward
In recent years, technologies like Stable Diffusion~\cite{stable-diffusion} and ControlNet~\cite{ControlNet} have revolutionized 2D AI-generated content (AIGC), making tasks like text-to-image synthesis, image editing, and style transfer more accessible and efficient. Concurrently, the potential of 3D AIGC has been recognized, allowing for the direct generation of 3D objects by integrating text and images, significantly reducing costs. Early optimization-based methods~\cite{Dream3d, Dreamfusion, Prolificdreamer}, where each object needs to be individually optimized, used multi-view stable diffusion~\cite{Zero-1-to-3, SyncDreamer, Bootstrap3D} which means generating images of an object from multiple perspectives by inputting an image from one perspective—to produce fine-grained objects but were slow, taking minutes to hours. However, feed-forward methods leveraging large-scale 3D asset datasets~\cite{Objaverse} and Transformer models now enable the creation of high-quality 3D objects in seconds. Despite progress in generation, advancements in 3D customization and editing, such as adding patterns or changing styles of 3D objects, are still scarce.

%%% Why 2 views.
In Feed-Forward Methods, although LRM~\cite{LRM} can generate high-quality 3D objects from a single view, it often lacks comprehensive details from other perspectives. In contrast, techniques like Instant3D~\cite{Instant3d} and LGM~\cite{LGM} use multi-view diffusion~\cite{MVDream, ImageDream} to generate images from four perspectives (front, back, left, and right) before reconstruction. While increasing the number of perspectives can capture more visual information, it also brings some challenges: managing multiple views simultaneously increases the complexity of editing tasks. For instance, if we want to change the color of a specific part of the object, it is difficult to precisely correspond the changes across all four images. To balance the richness of visual information and the ease of editing, we recommend prioritizing the front and back views. These views typically contain comprehensive information about the object and have minimal overlap, allowing them to be edited independently, thus simplifying operations and improving efficiency.

%%% Tailor3D
We propose an efficient and user-friendly 3D rapid editing framework, Tailor3D, which introduces a novel 3D editing way by leveraging advanced 2D image editing techniques. This framework delegates the generation and editing tasks to 2D image editing technologies and generates 3D objects through rapid 3D reconstruction, allowing users to iteratively refine the desired 3D objects through a combination of 2D editing and 3D reconstruction steps. The process is shown in \Cref{fig_teaser}: Assume the users have a front-view image of a dog. First, they edit the front view using image editing methods to generate space glasses and a dashboard seamlessly into the scene. Next, employing multi-view diffusion technology, they can generate a back view. Then they edit the back-view image with the image editing methods again to add the backpack. Finally, the edited front and back images are input into a Dual-sided LRM model to generate a 3D model of the space dog. The entire process allows for step-by-step editing and completes each step within seconds, providing great convenience for rapidly editing the required 3D objects. This step-by-step method provides more precise control than end-to-end editing, enabling specific adjustments to image textures before reconstruction. Additionally, separately editing front and back views allows for more detailed customization of the final 3D object.

%%% Dual-sided LRM
Our proposed Dual-sided LRM, used in the final step of Tailor3D, generates 3D objects by receiving front and back images. As shown in the lower part of \Cref{fig_teaser}, Having information from both sides allows for a more comprehensive understanding of the object, but it may lead to View inconsistency, referring to differences in geometry, color, and brightness in images taken from various angles and conditions, which can affect the quality of reconstruction.
We extends LRM’s capability from single-view to dual-view input, effectively handling inconsistencies between views. We introduce the LoRA Triplane Transformer~\cite{LoRA}, which fine-tunes the LRM model with minimal memory consumption on a small dataset of 20K images to generate triplane features for both front and back views. This approach efficiently produces accurate triplane features, providing a solid foundation for subsequent feature fusion. Instead of merely stitching 2D image features, we combine the 3D triplane features of both views within 3D space. By applying Viewpoint Cross-Attention on the triplane, we merge these features swiftly, enhancing the quality of the final 3D object. Additionally, we use data augmentation during training to further improve the model's robustness. Experimental results demonstrate that it excels in various 3D editing tasks, including geometric fill, texture synthesis, and style transfer.

% Summary
Our contributions can be summarized as follows:
\begin{enumerate}[\tiny$\bullet$]
\item We propose Tailor3D, a rapid 3D editing pipeline. By combining 2D image editing and rapid 3D reconstruction techniques, it significantly enhances the efficiency of 3D object editing.

\item Our Dual-sided LRM, combined with the LoRA Triplane Transformer, efficiently handles inconsistencies between front and back views, improving the overall reconstruction quality.

\item Tailor3D excels in various 3D editing and customization tasks, particularly in local 3D generative fill, overall style transfer, and style fusion for objects, showcasing immense practical utility.

\end{enumerate}

\section{Related Work}
\label{sec:2_Related_Work}
\textbf{Multi-view Diffusion for Objects.}
Utilizing a single front-view image, multi-view diffusion demonstrates remarkable capabilities in synthesizing images from alternate viewpoints of the object \cite{Zero-1-to-3, Zero123++, EscherNet, SyncDreamer, MVDiffusion++, MVDream, ImageDream}. These synthesized images are pivotal for subsequent stages of 3D object reconstruction to generate a mesh. 
% early work
Early efforts in this domain faced hurdles, particularly with small-scale training data and the imperative to ensure generalization performance \cite{3DiM, SparseFusion, GeNVS, Viewset_Diffusion, GPT-4V3D, Make-it-Real}.
% zero-1-to-3 and SyncDreamer
The improvement journey began with Zero-1-to-3 \cite{Zero-1-to-3} refining Stable Diffusion \cite{stable-diffusion} with the extrinsic camera parameters, marking a significant step in generalized multi-view diffusion. However, geometric consistency remained a challenge. SyncDreamer \cite{SyncDreamer} built upon Zero-1-to-3, introducing a 3D-aware feature attention mechanism for enhanced synchronization, yielding 16 highly coherent multi-view images.
% LRM series
Recent large models prefer using fewer overlapping canonical views (e.g., front, back, left, right) as inputs. This trend has led to the emergence of fixed-camera-parameter multi-view diffusion, simplifying training and enhancing multi-view consistency. For example, MVDream \cite{MVDream} and ImageDream \cite{ImageDream} efficiently generate these four views, while zero123++ \cite{Zero123++} extends this to six fixed views. Tailor3D improves practical utility by generating only the back image from the front, effectively addressing imperfect consistencies in diverse input scenarios.

\textbf{Large Model for 3D Reconstruction and Generation.}
% early methods
Early 3D generation methods initially focused on optimizing individual objects separately. SDS-based approaches~\cite{Dreamfusion, Dream3d, Magic3d, Realfusion, score, Dreambooth3d, Fantasia3d, Make-it-3d, Prolificdreamer, Hifa, Luciddreamer} utilized multi-view images from Zero-1-to-3 for this purpose. Subsequently, Diffusion + Reconstruction methods \cite{One-2-3-45, One-2-3-45++, Sculpt3d, Wonder3d} expanded on SyncDreamer to optimize higher-consistency multi-view images.
% LRM series
With the Large Reconstruction Model (LRM) scaling up in data and model size, it rapidly generates high-quality NeRF from single images in under 5s. This led to a shift where 2D methods handled generation tasks, and LRM managed 3D reconstruction. Consequently, 3D stable diffusion methods with fewer views, like MVDream \cite{MVDream}, became preferred. For instance, Instant3D \cite{Instant3d} uses 2D stable diffusion for four-view generation followed by LRM-like reconstruction. Similarly, LGM \cite{LGM} and GRM \cite{GRM} use Gaussian Splatting for reconstruction. For extensive 3D editing, we reduce perspectives to front and back, requiring lower consistency.

\textbf{3D Object Editing.}
% traditional way
In 3D object domain, "customized editing" involves shape alterations, pattern addition, and texture application under user control. Traditional methods include explicit geometric representation editing, such as mesh deformation \cite{Revisit, laplacianmesh, ARAP}, proxy-driven deformation \cite{fast, joint, free, neural-cages, mesh-based, efficient}, and data-driven deformation \cite{sparse, efficient}, which utilize prior shapes for realistic outcomes.
% NeRF way
Over time, editing has moved towards implicit radiance fields \cite{Neuroskinning, variational, Rignet}, especially on NeRFs \cite{EditingNeRF, ObjectNeRF, Nerf-editing}. Earlier works focused on specific objects or scenes, lacking generalization~\cite{GPT4Point}.
% 3D AIGC
In the 3D-AIGC era, 3D editing has evolved towards 2D image editing, reconstructed to generate new 3D objects \cite{Control3d, MVEdit}. MVEdit \cite{MVEdit} denoises multi-view images and outputs high-quality textured meshes. However, its inference process takes 2-5 minutes, lacking real-time editing. In contrast, Tailor3D uses dual-side LRM to process inputs from both object sides, completing each editing step within seconds, enabling interactive 3D object editing.

\section{Methodology}
\label{sec:3_Methodology}
\begin{figure}[!t]
  \vspace{-5mm}
  \centering
  \includegraphics[width=0.99\textwidth]{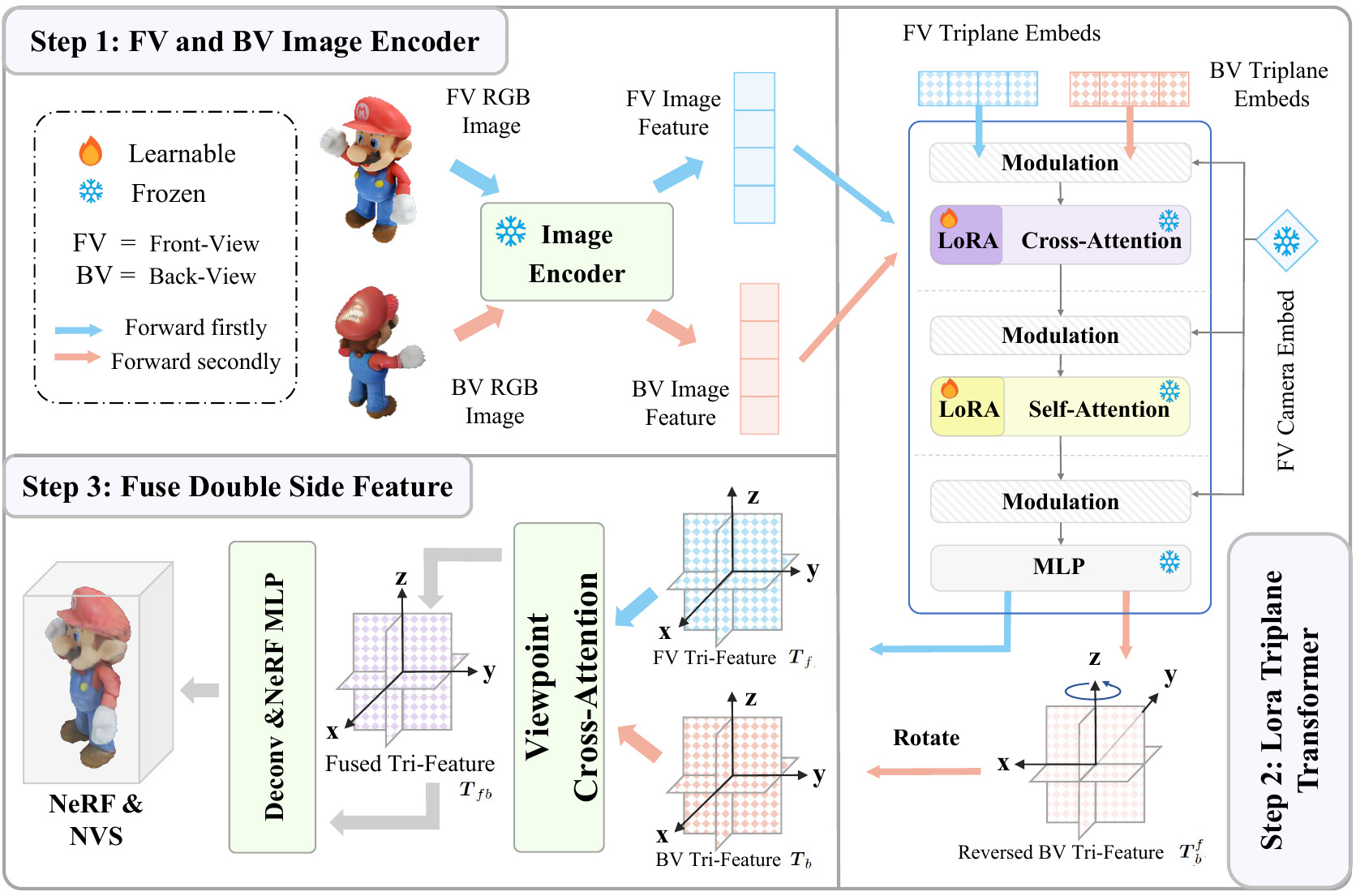}
  \caption{\textbf{Model Architecture of Dual-sided LRM}. We start with front and back view images. Then, using LoRA Triplane Transformer, we obtain front and back triplanes. Finally, we ‘tailor’ the two triplane features through rotation and Viewpoint Cross-Attention to obtain the 3D object.}
  \label{fig_arch}
  \vspace{-3mm}
\end{figure}

In this section, we present the pipeline and model architecture of Tailor3D. Firstly, we introduce the Large Reconstruction Model (LRM) and multi-view diffusion in \Cref{sec_3.1:Preliminaries}. Next, in \Cref{sec_3.2:The Pipeline of Tailor3D}, we outline Tailor3D's process, illustrating 2D editing and rapid reconstruction into 3D objects. In \Cref{sec_3.3:Dual-sided LRM: how to accept inconsistent views}, we delve into the Dual-sided LRM, accommodating inputs from imperfect consistent front and back views. We explain how the LoRA Triplane Transformer reduces memory usage and Viewpoint Cross-Attention to fuse 3D Triplanes from front and back views.

\subsection{Preliminaries}
\label{sec_3.1:Preliminaries}
\textbf{Large Reconstruction Model (LRM). }
LRM enables direct single-view to 3D reconstruction. The input image $\boldsymbol{I}$ is encoded by an image encoder, producing patch-wise feature tokens $\boldsymbol{F} \in {\mathbb{R}^{N \times d_E}}$, where $N$ is the number of image feature patches and $d_E$ is the dimension of the image encoder. Initial learnable positional embeddings for the triplane are defined as $\boldsymbol{f}^\mathit{init}$ and engage in cross-attention with the image features $\boldsymbol{F}$. They are modulated by the corresponding camera extrinsic parameters $\boldsymbol{E}$ to generate the triplane feature map $\boldsymbol{T}$.
\vspace{-0.2cm}
\begin{align}
     \boldsymbol{T} &= (\boldsymbol{T}_{{xy}},\boldsymbol{T}_{{yz}},\boldsymbol{T}_{{xz}}) = \textsc{Tri-Former}(\boldsymbol{f}^\mathit{init}, \ \boldsymbol{F}, \ \boldsymbol{E}) .
\end{align} 
Here, $\boldsymbol{f}^\mathit{init}\in (3{\times}32{\times}32){\times}{d_D}$, where $d_D$ is the hidden dimension of the transformer decoder. $\textsc{Tri-Former}$ incorporates self-attention, cross-attention, and modulation. The resultant triplane feature map $\boldsymbol{T} \in (3{\times}64{\times}64){\times}{d_T} $ comprises three planes: $\boldsymbol{T}_{X\!Y}$, $\boldsymbol{T}_{Y\!Z}$, and $\boldsymbol{T}_{X\!Z}$. Resolution increases from $32{\times}32$ to $64{\times}64$ via deconvolutional layers. Finally, it undergoes $\mathrm{MLP}^\mathit{nerf}$ for color and density derivation in NeRF rendering.

\textbf{2D and Multi-view Diffusion. }
The diffusion model iteratively denoises pure noise $x_{T}\sim \mathcal{N}(\mathbf{0}, \mathbf{I})$ over $T$ steps to yield clean data $x_0$, optimizing towards the gradient direction of the log probability distribution of the data, $\nabla_{\mathbf{x_t}}\log p(\mathbf{x_t})$. At step $t$, given the noisy input $x_t$, a neural network $\epsilon_\phi$ with parameters $\phi$ predicts the noise $\epsilon$.
\begin{equation}
\label{eq:diff}
\mathcal{L}_{\mathit{diff}}(\phi, x) = \mathbb{E}_{t, \epsilon} [\parallel\epsilon_\phi(x_t,t) - \epsilon \parallel_2^2] .
\end{equation}
Multi-view diffusion generates images from specific objects based on current and desired viewpoints. By providing current image $\boldsymbol{I}$, extrinsic camera parameters $\boldsymbol{E} \in {4{\times}4}$, alongside desired parameters camera 
$\boldsymbol{E}_{o}$, multi-view diffusion generates the image $\boldsymbol{I}_{o}$ for the desired viewpoint. In our pipeline, we utilize multi-view diffusion to generate the back image based on the front.

\subsection{The Pipeline of Tailor3D}
\label{sec_3.2:The Pipeline of Tailor3D}
This section outlines Tailor3D's pipeline, as shown in the lower part of \Cref{fig_teaser}. It begins with a front-facing image $\boldsymbol{I}_{f}$ of an object. Initially, image editing and style transfer are applied to create $\boldsymbol{I}_{f}^{\prime}$. Next, multi-view diffusion methods like Zero-1-to-3~\cite{Zero-1-to-3} generate the corresponding back image $\boldsymbol{I}_{b}$, which is then edited to get $\boldsymbol{I}_{b}^{\prime}$. Finally, both $\boldsymbol{I}_{f}^{\prime}$ and $\boldsymbol{I}_{b}^{\prime}$ are input into Dual-sided LRM to obtain the final 3D object. Tailor3D offers various choices and potential variations. Original images $\boldsymbol{I}_{f}$ and $\boldsymbol{I}_{b}$ can be directly input into Dual-sided LRM for rapid reconstruction of the 3D object. Additionally, the back image $\boldsymbol{I}_{b}$ can be generated not only through Zero-1-to-3 but also through photography or direct provision. We will further elaborate on downstream tasks in the experimental section. The flexibility of Tailor3D arises from improved choices at each step and the robustness of our model, Dual-sided LRM, in handling imperfect consistency between front and back image inputs.

\subsection{Dual-sided LRM: How to Accept Imperfect Consistent Views}
\label{sec_3.3:Dual-sided LRM: how to accept inconsistent views}
In \Cref{sec_3.2:The Pipeline of Tailor3D}, our focus is on acquiring the edited front image $\boldsymbol{I}_{f}^{\prime}$ and back image $\boldsymbol{I}_{b}^{\prime}$ for an object. However, these images may exhibit imperfect consistency: They might not directly face the object, and their relationship can vary. Therefore, we need a reconstruction model capable of handling imperfectly consistent input images from both views to generate 3D objects. We select two views instead of four to reduce inconsistency pressure on editing and reconstruction. We explicitly merge two triplane features in the 3D domain, aiming to resolve the inconsistency issue intuitively.

\begin{figure}[!t]
  \centering
  \includegraphics[width=0.99\textwidth]{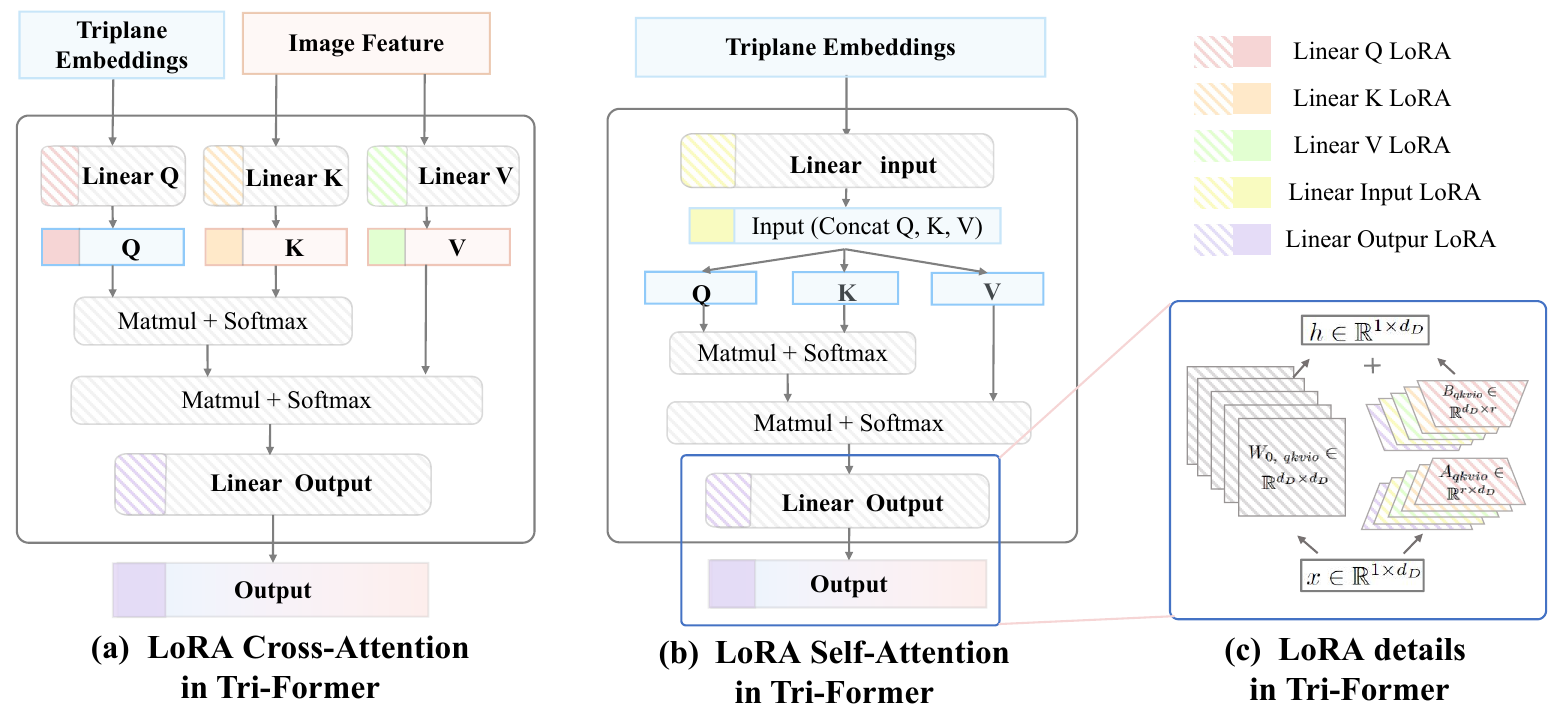}
  \caption{\textbf{LoRA Triplane Transformer}. (a) For Cross-Attention, we use the LoRA structure to replace the connection layers of $qkv$ and $output$. (b) For Self-Attention, we replace the connection layers of $input$ and $output$. Details of the LoRA are shown in (c).}
  \label{fig_lora}
  \vspace{-4mm}
\end{figure}

\textbf{LoRA Triplane Transformer. }
% Triplane transformer for LRM
When employing pre-trained LRM parameters~\cite{LRM}, our goal is to minimize memory usage. In LRM, the single view feature $\boldsymbol{F}_{f}^{\prime}$ is processed by a triplane transformer serving as a decoder to generate triplane NeRF features $\boldsymbol{T}_{f}$. This component facilitates mapping from a single view to 3D, enabling the model to understand diverse object shapes and infer object information effectively.
% LoRA triplane transformer
To minimize memory usage, we integrate the LoRA structure into the triplane transformer, as depicted in \Cref{fig_lora}. For self-attention, where $qkv$ is generated by shared linear layers, we replace all input and output linear layers with LoRA structures~\cite{LoRA}. For cross-attention, where $qkv$ is generated by different linear layers, we replace all $qkv$ and output linear layers with LoRA structures. Specific details are as follows:
\begin{equation}
h^{i} = W_{0}^{i}x + \Delta W_{tp}^{i}x = W_{0}^{i}x + B_{tp}^{i}A_{tp}^{i}x .
\end{equation}
Here, $i$ denotes the $i$-th Transformer layer. For self-attention, $tp$ represents the linear projection for $input$ and $output$. For cross-attention, $tp$ denotes the linear projections for $q, k, v$, and $output$.

As shown in \Cref{fig_arch}, LRM generates the triplane feature $\boldsymbol{T}_{f}$ for the front view from features $\boldsymbol{F}_{f}^{\prime}$ and camera parameters $\boldsymbol{E}_{f}$. Similarly, for the back view features $\boldsymbol{F}_{b}^{\prime}$, we use the camera parameters $\boldsymbol{E}_{f}$ of the front view to obtain the triplane feature $\boldsymbol{T}_{b}^{f}$ for the back view through the LoRA triplane transformer, as expressed by the following equation:
\begin{align}
\boldsymbol{T}_{f} / \boldsymbol{T}_{b}^{f} = \textsc{Tri-Former}_{\texttt{LoRA}}(\boldsymbol{f}^\mathit{init}, \ \boldsymbol{F}_{f}^{\prime}/\boldsymbol{F}_{b}^{\prime}, \ \boldsymbol{E}_{f}) .
\end{align}
Here $\boldsymbol{T}_{b}^{f}$, the triplane feature for the back view obtained using the front view's camera parameters, cannot be directly merged with $\boldsymbol{T}_{f}$. We will address this and the inconsistency between the front and back view angles in the next section.

\textbf{Fuse Double Side Feature. }
To merge the two triplane features $\boldsymbol{T}_{f}$ and $\boldsymbol{T}_{b}^{f}$, we first horizontally flip $\boldsymbol{T}_{b}^{f}$ by 180 degrees around the z-axis to obtain $\boldsymbol{T}_{b}$.
% Viewpoint Cross-Attention
Due to inconsistency between the front and back views, direct alignment or addition of the triplane features isn't feasible. Leveraging the triplane representation, we apply Viewpoint Cross-Attention to each plane individually. We use $\boldsymbol{T}_{f}$ as the query and $\boldsymbol{T}_{b}$ as the key and value to incorporate missing information from the backside. We adopt a window-based attention structure, with a window size set to 7, significantly reducing memory consumption. This yields the final $\boldsymbol{T}_{fb}$, encapsulating information from both views. Data augmentation further bolsters robustness to inconsistency, with back view images undergoing scaling, rotation, and translation, each with a $10\%$ probability.

Finally, the Triplane-NeRF formulation utilizes $\mathrm{MLP^\mathit{nerf}}$ to derive NeRF color and density parameters for volume rendering. Supervision includes $V$ views, comprising the front, back and $(V-2)$ randomly chosen side views. For a specific view $v$, the loss function for synthesizing the prediction $\hat{x}_{v}$ and the ground truth $\boldsymbol{x}^\mathit{GT}_{v}$ for new view composition is formulated as follows:
\begin{align}
\mathcal{L}(\boldsymbol{x}) &= \frac{1}{V} {\sum_{v=1}^{V}} \left( {\lambda_1}\small\mathcal{L}_\mathrm{MSE}(\boldsymbol{\hat{x}}_{v},\boldsymbol{x}^\mathit{GT}_{v})+{\lambda_2}\mathcal{L}_\mathrm{LPIPS}(\boldsymbol{\hat{x}}_{v},\boldsymbol{x}^\mathit{GT}_{v}) + {\lambda_3}\mathcal{L}_\mathrm{TV}(\boldsymbol{\hat{x}}_{v},\boldsymbol{x}^\mathit{GT}_{v})\right) .
\label{eqn:loss}
\end{align}
$\mathcal{L}_\mathrm{MSE}$ denotes the normalized pixel-wise L2 loss, $\mathcal{L}_\mathrm{LPIPS}$ is perceptual image patch similarity. $\mathcal{L}_\mathrm{TV}$ is the total variation loss to prevent noise in the image. Weight coefficients $\lambda_1, \lambda_2, \lambda_3$ are applied.

\section{Experiments}
\label{sec:4_Experiments}
This section explores the experimental aspects. We begin with insights into the Gobjaverse-LVIS~\cite{Gobjaverse, LVIS} dataset in \Cref{sec:4.1_Dataset: Gobjaverse-LVIS}. In \Cref{sec:4.2_Implementation Details}, we delve into various implementation details, including model architecture parameters, camera adjustments, and training/testing processes. In \Cref{sec:4.3_Experiment Results}, we present experimental results. We showcase Tailor3D's versatility across different tasks and conduct ablation studies on key modules like the LoRA Triplane Transformer and fusion techniques.

\subsection{Dataset: Gobjaverse-LVIS}
\label{sec:4.1_Dataset: Gobjaverse-LVIS}
LRM pre-trained weights~\cite{LRM, OpenLRM} are trained on the Objaverse~\cite{Objaverse} and MVImgNet~\cite{MVImgNet} datasets, containing 730K objects, normalized to a cube of size $[-1, 1]^{3}$ and rendered from 32 random viewpoints at a resolution of $512 \times 512$ pixels. For fine-tuning, the Gobjaverse-LVIS~\cite{Gobjaverse, LVIS} dataset comprises 22K high-quality 3D rendered objects, selected from G-buffer Objaverse and LVIS datasets. Gobjaverse includes 280K 3D objects captured from various viewpoints. During training, matching front and back views with identical elevation are used. Rendering supervision includes fixed front and back viewpoints, along with $(V-2)$ randomly selected side views for new view synthesis. The combined Gobjaverse-LVIS dataset consists of 22K objects, ensuring higher quality. % Additional details are provided in \Cref{sec:D_Additional_Experiments}.

\subsection{Implementation Details}
\label{sec:4.2_Implementation Details}
We use the network architecture from the pre-trained LRM model. The image encoder is based on DINOv2's ViT-B/16 model~\cite{DINOv2}, operating at a resolution of $384{\times}384$. The image features have a dimensionality of 768. The triplane transformer decoder consists of 16 layers with 16 transformer heads, featuring positional embeddings of dimensionality 1024 and triplanes with dimensionality 80. $\mathrm{MLP}^\mathit{nerf}$ comprises 10 layers. We set the LoRA rank to 4 for the LoRA Triplane Transformer. During neural rendering, we sample 128 points along each ray and produce images at a resolution of $128{\times}128$. For camera normalization, we align with LRM standards, positioning the camera at $[0, -2, 0]$ relative to the object center. This ensures the object's z-axis is upward, and the front view corresponds to the negative y-axis. External rendering parameters are normalized relative to the reference view. We train for 10 epochs on 8 A100 GPUs with a batch size of 16, taking about 6 hours. The loss function coefficients are $\lambda_1 = \lambda_2 = \lambda_3 = 1.0$. We use the AdamW optimizer with a learning rate of $3{\times}10^{-4}$ and a cosine schedule. During inference, we query a resolution of $384{\times}384{\times}384$ points from the reconstructed triplane-NeRF, completing it in less than 5 seconds.

\subsection{Experiment Results}
\label{sec:4.3_Experiment Results}
In \Cref{sec:4.3.1_Tailor3D Applications}, we showcased Tailor3D's capabilities in 3D generation, covering geometric object fill, texture synthesis, and style transfer. In \Cref{sec:4.3.2_Compared to Existing 3D Generations}, we compared our approach with existing techniques. In \Cref{sec:4.3.3_Ablation Study}, we performed ablation experiments to validate each module of Tailor3D.

\subsubsection{Tailor3D Applications}
\label{sec:4.3.1_Tailor3D Applications}
We showcase its versatility in 3D Generative Geometry / Pattern Fill, encompassing local geometric shape and texture pattern filling. We highlight its style transfer and fusion capabilities, allowing for operations like style transfer and blending two styles onto one object. Tailor3D enables users to edit both the front and back of objects, expanding editing possibilities for customized 3D objects.

\begin{figure}[!t]
  \vspace{-9mm}
  \centering
  \includegraphics[width=0.98\textwidth]{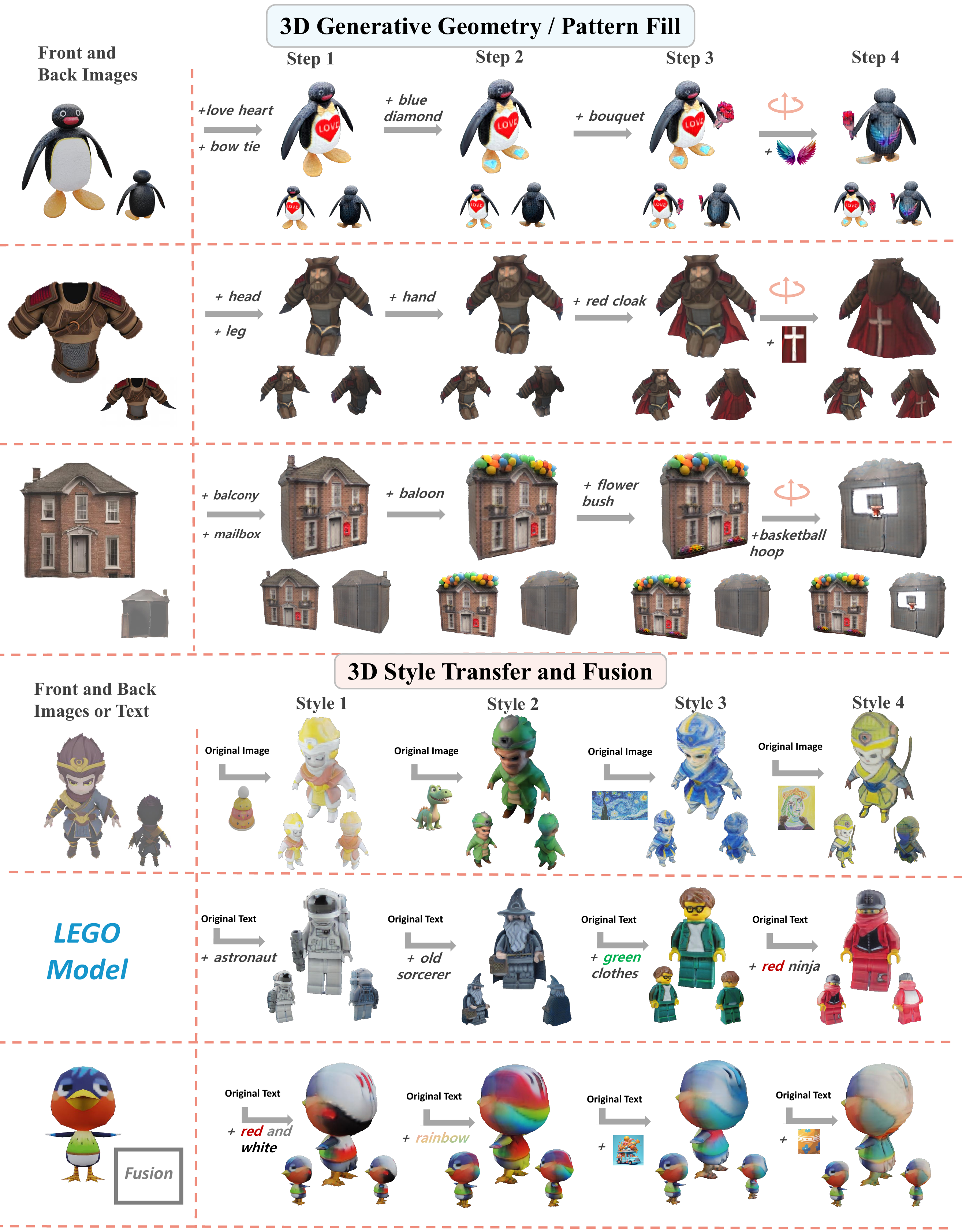}
  \caption{\textbf{3D Generative Fill and 3D Style Transfer.} It includes both Geometry Fill and Pattern Fill, allowing us to add or modify local geometric structures or texture patterns of 3D objects. Guidance can be provided through text or images as prompts. Additionally, we offer style images or textual guidance to transform 3D objects into desired styles. Ensuring the maintenance of IP integrity during disguise adds significant practical value to 3D tasks.}
  \label{fig_exp_fill_and_style}
  \vspace{-4mm}
\end{figure}

\textbf{3D Generative Geometry / Pattern Fill.} Here, we showcase Tailor3D's local 3D object filling ability, as depicted in \Cref{fig_exp_fill_and_style}. Demonstrating step-by-step object filling and editing through text or image prompts. In Row 2, starting from armor, we generate a medieval general by adding the head, hands, and cloak progressively. Row 3 illustrates additional object manipulation, including the addition of a mailbox, balloons, a flower bush, and a basketball hoop.

\textbf{3D Style Transfer and Fusion.} Tailor3D also demonstrates its transfer and fusion capabilities for various styles. Unlike previous approaches, Tailor3D ensures IP integrity while offering flexibility in specifying styles through images or text guidance. Notably, it leverages Midjourney for 2D image generation and editing. Additionally, Tailor3D enables the infusion of different styles onto both the front and back of objects, showcasing the effectiveness of the Dual-sided LRM's merging ability.

\subsubsection{Compare to Existing 3D Image-to-3D Generation Methods}
\label{sec:4.3.2_Compared to Existing 3D Generations}
\begin{figure}[!t]
  \vspace{-7mm}
  \centering
  \includegraphics[width=0.99\textwidth]{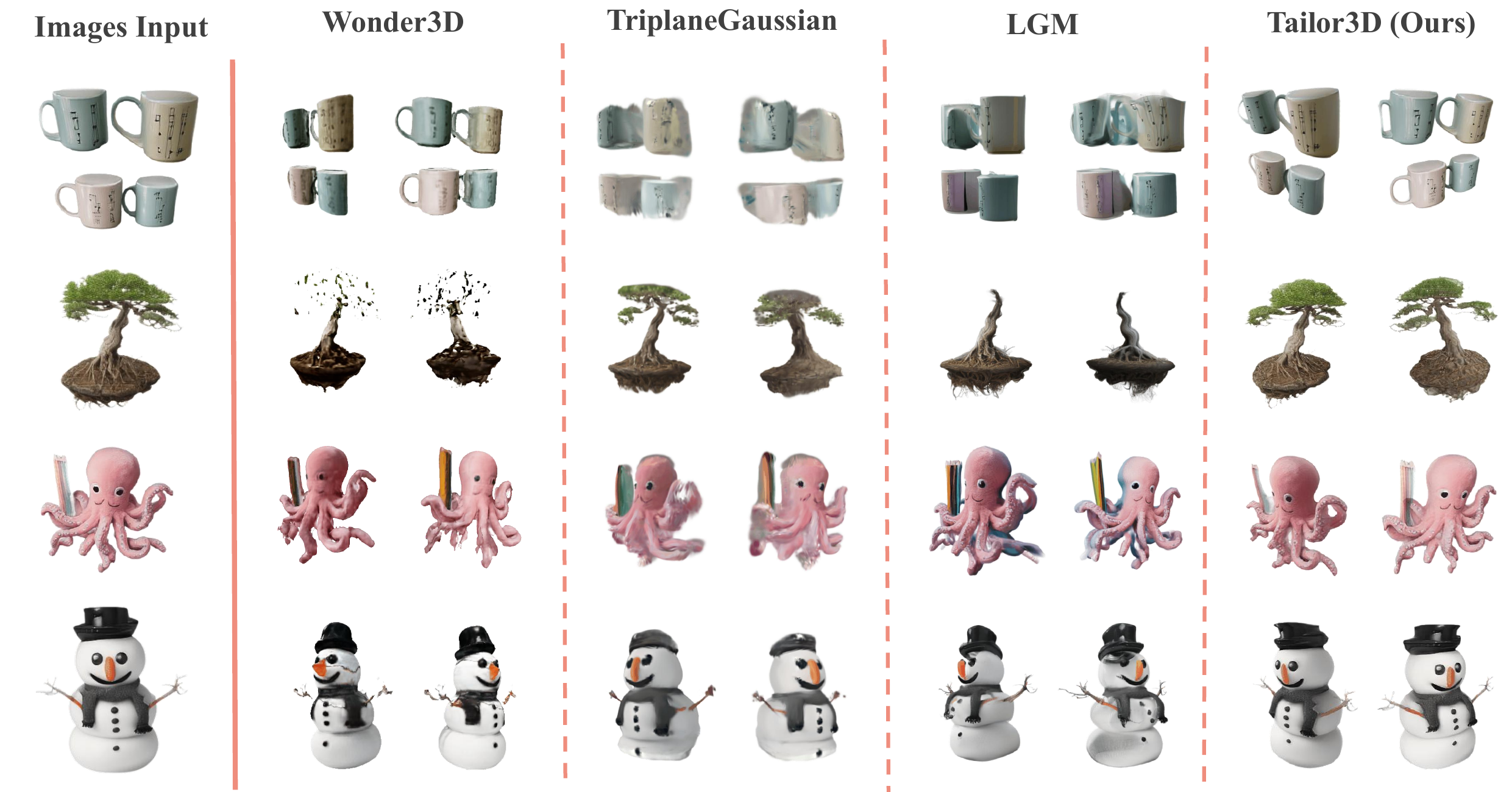}
  \caption{\textbf{Compare to Existing 3D Generation. }We compare single image-to-3D methods. Wonder3D and TriplaneGaussian have lower resolutions, while LGM often shows ghosting effects with complex textures. Our method, however, achieves superior experimental results.}
  \label{fig_exp_compare}
  \vspace{1.4mm}
\end{figure}

We compare our approach with Wonder3D~\cite{Wonder3d}, TriplaneGaussian~\cite{TriplaneGaussian}, and LGM~\cite{LGM} on a test set of 100 images generated by stable diffusion~\cite{stable-diffusion}. Qualitative results in \Cref{fig_exp_compare} demonstrate Tailor3D's capability to enhance backside information with Dual-sided LRM. Wonder3D and TriplaneGaussian struggle with complex objects, exhibiting lower overall quality. LGM, using Gaussian representation, suffers from ghosting effects and lacks detail in features like tree leaves. Quantitative results are provided in \Cref{tab:results_compared} alongside generation times, highlighting the practical value of our method.

\subsubsection{Ablation Study}
\label{sec:4.3.3_Ablation Study}
We perform an ablation study on the Dual-sided LRM, focusing on three aspects: the fusion of 3D features from both sides, the rank of the LoRA Transformer, and the extrinsic camera parameters of front and back images. Results are presented in \Cref{tab:Abalation Study}, using the same test set as in \Cref{sec:4.3.2_Compared to Existing 3D Generations}.

\textbf{The Way to Fuse Double Side Feature. }We use the Viewpoint Cross-Attention to fuse features from both the front and back sides. Additionally, we experiment with multiple layers of 2D convolutional layers and direct addition to merge Triplane features from both sides. Our results indicate that employing the Viewpoint Cross-Attention produces the best results.

\textbf{The Rank of LoRA Triplane Transformer. }We conduct ablation experiments on the rank of the LoRA Triplane Transformer, setting the rank to 2, 4, and 8, respectively. Our experimental results indicate that a rank of 4 achieves the best performance.

\textbf{Extrinsic Camera Parameters. }We apply the same front camera parameters $\boldsymbol{E}_{f}$ to both front and back images, rotating only the back triplane. Additionally, we experiment with separate camera parameters for front and back images, without rotation, by utilizing both the front and back camera extrinsics, denoted as $\boldsymbol{E}_{f}$ and $\boldsymbol{E}_{b}$, respectively. The results suggest that using front extrinsics alone yields accurate outcomes, as the LRM structure solely accepts front camera parameters.

In our ablation study, we found that Viewpoint Cross-Attention is more effective than convolutional networks for merging 3D features. A rank of 4 for the LoRA Triplane Transformer yields optimal results, while the LRM framework only accepts front-facing camera parameters.
% Through the ablation study, we found that both convolutional networks and Viewpoint Cross-Attention are effective in stitching together the two 3D features, with Viewpoint Cross-Attention yielding better results. Regarding the rank of the LoRA Triplane Transformer, we observed that this value has minimal impact on the overall results, but achieves optimal performance when the rank is set to 4. As for the camera parameters, we discovered that the LRM framework can only accept front-facing camera parameters.

\begin{table}[!t]
    \begin{minipage}{1\textwidth}
    \centering
        \tablestyle{5.8pt}{1.08}
        \begin{tabular}{lclllcccc}\toprule
\multicolumn{2}{c}{Compare with others.} &\multicolumn{3}{c}{Common Metrics} &\multicolumn{3}{c}{User Study $\uparrow$ \small{(0 to 100 score)} } \\\cmidrule(lr){3-5} \cmidrule(lr){6-8}
Methods & InF. Time. & LPIPS $\downarrow$ & SSIM $\uparrow$ & PSNR $\uparrow$ \ \ \  & Geometry & Texture & Overall\ \ \  \\\midrule
TriplaneGaussian\cite{TriplaneGaussian} & 20s & 0.2811 & 0.5635 & 14.89 & 56.3 & 54.5 & 62.3\\

Wonder3D~\cite{Wonder3d} & 3min & 0.2709 & 0.6485 & 16.23 & 73.3 & 76.3 & 79.2\\

LGM~\cite{LGM} & 5s & 0.2473 & 0.8423 & 19.02 & 79.3 & \textbf{85.2} & 83.2\\

Tailor3D (Ours) & 5s & \textbf{0.2345} & \textbf{0.8525} & \textbf{19.34} & \textbf{82.3} & 84.2 & \textbf{86.3}\\
\bottomrule
\end{tabular}
        \caption{\textbf{Comparison with Existing 3D Generation Methods.} We compare single image-to-3D methods, including common metrics and user studies. Results indicate that ours outperforms others.}\label{tab:results_compared}
        \vspace{-1mm}
    \end{minipage}
    \vspace{1.5mm}
    % \vspace{-0.5mm}
    \begin{minipage}{0.31\textwidth}
        \subcaption{\textbf{Way to Fuse Double Sides.}
    \vspace{1.5mm}}\label{subtab:aba_way_fuse}
        \vspace{-2mm}
        \centering
        \tablestyle{0pt}{1.08}
        \begin{tabular}{x{40pt}| x{25pt} x{30pt} x{30pt}}\toprule
Fuse Way & Score & SSIM$\uparrow$ & LPIPS$\downarrow$ \\\midrule
Add  & 76.3  & 0.7377 & 0.2938 \\
Conv2D & 84.2 & 0.8239 & 0.2443  \\
VP-CA$\dagger$ & \textbf{86.3} & \textbf{0.8525} & \textbf{0.2345}  \\
\bottomrule
\end{tabular}
    \end{minipage}
    \hspace{2mm}
    \begin{minipage}{0.31\textwidth}
        \centering
        \subcaption{\textbf{LoRA Transformer Rank.}\vspace{1.5mm}}\label{subtab:aba_rank_lora}
        \vspace{-2mm}
        \tablestyle{0pt}{1.08}
        \begin{tabular}{x{25pt}| x{25pt} x{30pt} x{30pt}}\toprule
Rank & Score & SSIM$\uparrow$ & LPIPS$\downarrow$ \\\midrule
2  & 79.2 & 0.7623 & 0.2877 \\
4 & \textbf{86.3} & \textbf{0.8525} & \textbf{0.2345} \\
8  & 82.2  & 0.7902 & 0.2535  \\
\bottomrule
\end{tabular}
    \end{minipage}
    \hspace{1mm}
    \begin{minipage}{0.31\textwidth}
        \centering
        \subcaption{\textbf{Two Camera Extrinsics.}\vspace{1.5mm}}\label{subtab:aba_cam_extrinsic}
        \vspace{-2mm}
        \tablestyle{0pt}{1.08}
        \begin{tabular}{x{45pt}| x{25pt} x{30pt} x{30pt}}\toprule
Cam Ext. $*$& \multirow{1}{*}{Score} & \multirow{1}{*}{SSIM$\uparrow$} & \multirow{1}{*}{LPIPS$\downarrow$} \\\midrule
$\boldsymbol{E}_{b}$ + $\boldsymbol{E}_{b}$  & 60.5 & 0.6288 & 0.3944 \\
$\boldsymbol{E}_{f}$ + $\boldsymbol{E}_{b}$ & 33.4  & 0.3523 & 0.5653  \\
$\boldsymbol{E}_{f}$ + $\boldsymbol{E}_{f}$ & \textbf{86.3} & \textbf{0.8525} & \textbf{0.2345}  \\
\bottomrule
\end{tabular}
    \end{minipage}
    \vspace{1.6mm}
    \caption{\textbf{Abalation Study.} We conducted ablation regarding the fusion method for both sides, the rank of the LoRA Triplane Transformer, and the extrinsic camera parameters. $\dagger$: VP-CA means Viewpoint Cross-Attention. $*$: The first is the front-view extrinsic and the second is for the back view.}\label{tab:Abalation Study}
    \vspace{-2mm}
\end{table}

\section{Limitation and Conclusion}
\label{sec:5_Limitation_and_Conclusion}
\vspace{-3mm}
In this paper, we introduce Tailor3D, which swiftly creates customized 3D assets using editable dual-side images, akin to a tailor's approach. By leveraging 2D image editing techniques and rapid 3D reconstruction, Tailor3D allows users to iteratively refine objects. Our Dual-sided LRM and LoRA Triplane Transformer act as 'tailors,' seamlessly stitching together front and back views to handle inconsistencies and improve reconstruction quality. Experimental results validate Tailor3D's effectiveness in tasks like 3D generative fill and style customization. It offers a user-friendly, cost-efficient solution for rapid 3D editing, applicable in animation, game development, and beyond, streamlining production and democratizing content creation. 

\noindent{\textbf{Limitation and Future Direction.}}
However, relying solely on front and back views for object reconstruction may encounter challenges with objects of certain thicknesses. Additionally, the generated 3D object meshes may have lower resolutions, and the addition of geometric features may not significantly alter the mesh. We will further investigate methods to address the generation and reconstruction of objects with thicker side profiles in future work, aiming to enhance the quality and resolution of the meshes.

\newpage
{
\small
\bibliography{main}
\bibliographystyle{plain}
}

\newpage
\appendix

\section{Additional Introduction}
\label{sec:A_Additional_Introduction}
We first introduce additional background information in the supplementary materials in \Cref{sec:B_Additional_Related_Work}. We first divided 3D Reconstruction into three categories and introduced the LRM~\cite{LRM} family. In \Cref{sec:C_Additional_Methodology}, we presented additional details regarding the methodology and implementation of experiments. We emphasize the differences between our training configuration and the original LRM and provide further insights into the Gobjaverse~\cite{Gobjaverse} dataset. In \Cref{sec:D_Additional_Experiments}, we primarily showcase the additional experimental content we have supplemented. We first present additional examples of Tailor, followed by comparisons with more multi-view reconstruction methods. In \Cref{sec:E_Broader_Impacts}, we discuss the broader social impact of our process.

\section{Additional Related Work}
\label{sec:B_Additional_Related_Work}
This section categorizes 3D reconstruction into single-view reconstruction, multi-view reconstruction, and the recently popular normal-view reconstruction. We then delve into the benefits of employing double-sided information for canonical-view reconstruction in \cref{sec:B_1_Single Multi and Normal-view Reconstruction}. Following that, we introduce articles from the LRM family~\cite{LRM, Instant3d, PF-LRM, DMV3D} in \cref{sec:B_2_Introduction to LRM Family}, discussing various variants of this universal reconstruction framework.

\subsection{Single, Multi and Canonical-view Reconstruction}
\label{sec:B_1_Single Multi and Normal-view Reconstruction}
Firstly, we delineate several types of reconstruction. Single-view reconstruction involves generating a 3D mesh of an object from a single viewpoint image (typically the front view). On the other hand, multi-view reconstruction typically involves multiple viewpoint images of an object along with corresponding camera extrinsic (often 20-100 views), aiming to reconstruct a 3D object. A landmark method in this domain is NeRF, which utilizes MLPs for novel view synthesis or 3D reconstruction. However, NeRF-based methods suffer from the need for individual optimization for each object, resulting in long reconstruction times, sometimes reaching 1-2 hours. Early 3D generation methods which use multi-view diffusion for generating multiple views of an object and subsequent reconstruction~\cite{One-2-3-45, Wonder3d}, also face long reconstruction times.

The development trajectory of NeRF involves the need for increasingly fewer viewpoints for reconstruction, fewer camera parameters, and faster reconstruction speeds. However, these methods still require individual optimization for each object. In contrast, LRM serves as a universal reconstruction model. As the model and dataset sizes reach a particular scale, reconstruction models become universal, eliminating the need for individual optimization of objects to be reconstructed. Within this universal framework emerges a reconstruction method known as canonical-view reconstruction, which uses fixed faces for reconstruction, typically the front, back, left, and right faces, referred to as 4-canonical-view reconstruction. Instant3D~\cite{Instant3d}, TriplaneGaussian~\cite{TriplaneGaussian}, and LGM~\cite{LGM} all employ this reconstruction method. However, the challenge with using the front, back, left, and right faces lies in effective editing, as it is difficult to edit all four faces simultaneously. Tailor3D adopts \textbf{Dual-Canonical-view Reconstruction}, utilizing only the front and back faces with fewer overlaps, facilitating user editing. Here, we emphasize that multi-view reconstruction requires optimization for individual objects, whereas canonical-view reconstruction is built upon a general reconstruction framework.

\begin{figure}[t]
  \centering
  \includegraphics[width=0.84\textwidth]{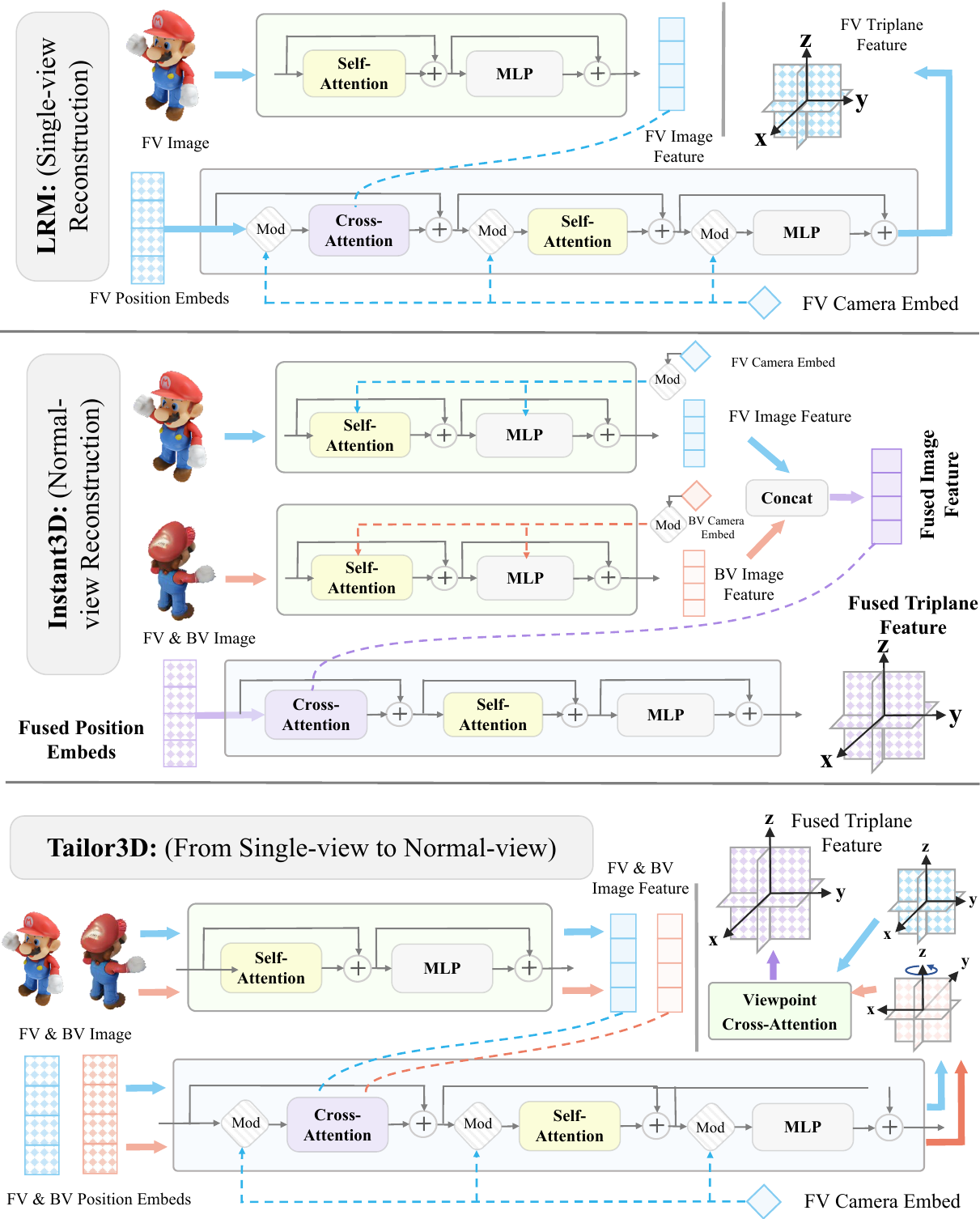}
  \vspace{-10pt}
  \caption{Model architectures of LRM, Instant3D and Tailor3D.}
  \label{fig_background}
\end{figure} 

\subsection{Introduction to LRM Family}
\label{sec:B_2_Introduction to LRM Family}
As mentioned earlier, early 3D generation methods utilized multi-view diffusion to generate additional viewpoints from a single image and optimized the multi-view reconstruction of a 3D object based on these views which need \textbf{several minutes}. The LRM family, serving as a series of Feed-Forward Methods, directly generates 3D meshes without the need for synthesizing multiple viewpoint images or training and adapting to models like NeRF \textbf{within only several seconds}. It represents a universal reconstruction framework. As illustrated in \Cref{fig_background}, LRM is a universal framework for single-view reconstruction. That is, a single image can directly generate a 3D mesh. The fundamental concept involves predefining the feature map of Triplane NeRF and then performing cross-attention with 2D images and their corresponding camera parameters. The resulting feature map can directly provide novel views of images or even the entire 3D mesh in the format of Triplane NeRF.

Building upon this foundation, Instant3D~\cite{Instant3d} addresses normal 4-canonical-view reconstruction. It involves two stages: first, utilizing a 2D diffusion model to obtain front, back, left, and right images of an object from text prompts; second, reconstructing the 3D object from these four viewpoints. PF-LRM~\cite{PF-LRM} focuses on pose-free sparse multi-view reconstruction, enabling the generation of a 3D object from three images taken from arbitrary viewpoints without corresponding camera extrinsics. However, its framework complexity arises from the supervision involving PnP and various geometric theories. DMV3D~\cite{DMV3D}, an extension of Instant3D, introduces a denoising process, resulting in a denoised multi-view diffusion framework. Unfortunately, these methods have not been open-sourced yet, with only the OpenLRM~\cite{OpenLRM} codebase providing the inference code for LRM.

LRM and Instant3D can be regarded as methods corresponding to single-view and 4-canonical-view reconstruction, respectively. However, their handling of camera parameters differs. As shown in \cref{fig_background}, LRM adjusts camera parameters with triplane features in the triplane transformer decoder. In practice, the external camera parameters are fixed, meaning the camera is positioned at $[0, -2m, 0]$ and oriented to look directly at the object along the positive y-axis. Hence, LRM can only accept the camera parameters of the front view, as demonstrated in \Cref{subtab:aba_cam_extrinsic}. In contrast, Instant3D places the modulation of the camera within the image encoder. After obtaining image features from four views, these features are concatenated and passed through the triplane transformer decoder. This approach involves merging the features from multiple viewpoints at the 2D image feature level. However, this approach is not a natural transition from single-view to canonical-view reconstruction. We choose to utilize Viewpoint Cross-Attention to fuse the 3D triplane features of the front and back views. This allows us to easily extend single-view reconstruction to dual(4)-canonical-view reconstruction using only the pre-trained weights from the single-view reconstruction. Furthermore, only training the Viewpoint Cross-Attention is necessary to minimize costs.

\section{Additional Methodology}
\label{sec:C_Additional_Methodology}
\begin{figure}[t]
  \centering
  \includegraphics[width=0.90\textwidth]{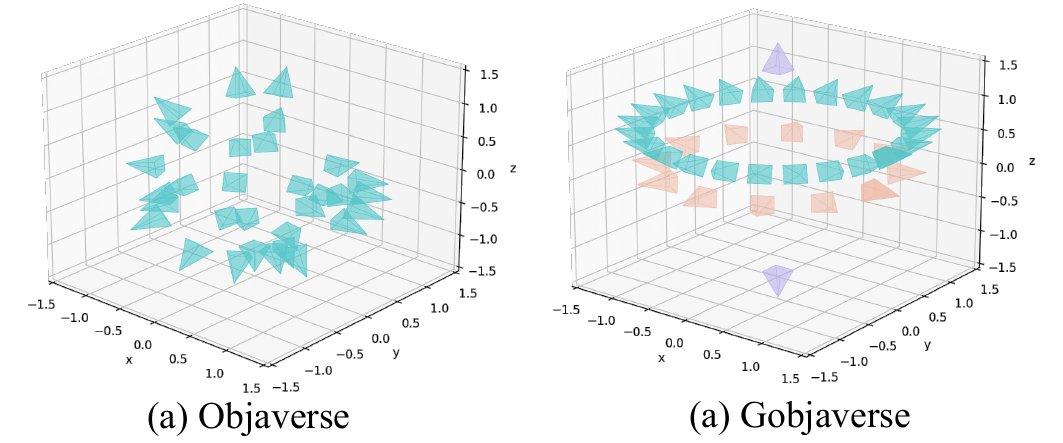}
  \vspace{-1mm}
  \caption{Rendering perspectives in Objaverse and Gobjaverse.}
  \label{fig_gobjaverse}
  \vspace{-6mm}
\end{figure} 

In this section, we delve into the training and experimental aspects. In \cref{sec:C_1_Training Settings}, we outline our training setup, leveraging the LRM model from the OpenLRM codebase~\cite{OpenLRM}, and delineate the variations in parameter quantities compared to the original LRM. In \cref{sec:C_2_Dataset: GObjaverse}, we offer a detailed overview of the viewpoint rendering in the Gobjaverse dataset~\cite{Gobjaverse} utilized in our study. We achieved satisfactory results with a relatively small dataset size by utilizing meticulously crafted artificial rendering data boasting high-quality textures and excellent consistency (22K).

\subsection{Training Settings}
\label{sec:C_1_Training Settings}
Here, we focus on describing our training details. First, we utilized the OpenLRM codebase as the basis for our LRM implementation. The original resolution is 512, but we used 256. The dimensionality of the triplane feature map, which was initially 80, was reduced to 40. Other model parameters remain unchanged, such as the dimensionality of camera embeddings (1024) and triplane transformer (1024). We used 96 rendering sample rays. For training parameters, the learning rate was set to $3e-4$, with a weight decay of 0.05. We employed a cosine scheduler. The total batch size was set to 16 (across 8 A100 GPUs), and we trained for a total of 20 epochs.

\subsection{Dataset: Gobjaverse}
\label{sec:C_2_Dataset: GObjaverse}
We utilized the Gobjaverse dataset~\cite{Gobjaverse}, an enhanced version of the Objaverse dataset with higher-quality rendering. Unlike Objaverse, which renders a single object with randomly positioned cameras spherically, Gobjaverse performs orbit rendering around an object, capturing two orbits shown in \Cref{fig_gobjaverse}. In the higher-elevation orbit, 24 views at equal intervals are represented in cyan. In the lower-elevation orbit, 12 views at equal intervals are represented in red. Additionally, two views captured from the top and bottom are represented in purple.

We excluded the two views captured from the top and bottom during our training process. This allowed our training data to provide input from both the front and back sides of the objects. It is worth noting that the opposite directions are only along the x-axis and y-axis. In the z-axis direction, they have the same elevation angle rather than being utterly symmetric across the center. This approach differs from methods like Instant3D and LGM~\cite{LGM}, which use techniques similar to MVDream~\cite{MVDream} to generate 4 views of an object using 2D diffusion. Gobjaverse offers higher consistency, resulting in higher data quality, which facilitates the fusion of features from the front and back directions.

\subsection{Testset: 100 Images from Stable Diffusion}
\label{sec:C_3_Testset}
Our quantitative test set and a portion of the qualitative test set consist of 100 objects generated by Stable Diffusion, with the background removed. Here, we present partial examples using two images, while the remaining qualitative examples may come from the use of Midjourney for generation. Our test set covers various objects and micro-scenes such as animals, humans, plants, and landscapes, enabling a comprehensive assessment of the quality of the generation models. Additionally, all our models comply with copyright and related regulations.

\begin{figure}[t!]
  \centering
  \includegraphics[width=0.99\textwidth]{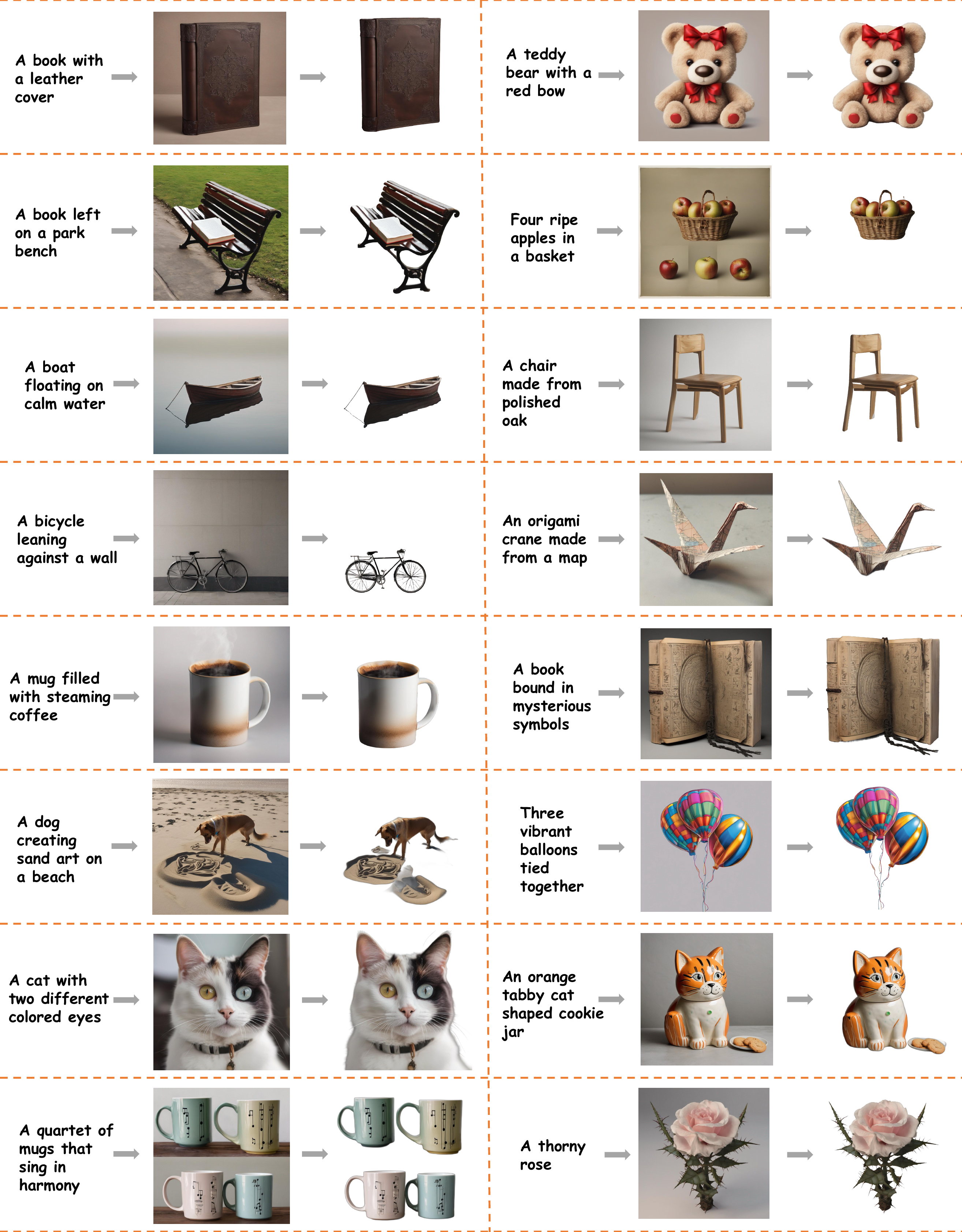}
  \vspace{-1mm}
  \caption{Testset: 100 3D Assets from Stable Diffusion (1).}
  \label{fig_100sd_1}
  \vspace{3mm}
\end{figure} 

\begin{figure}[t!]
  \centering
  \includegraphics[width=0.99\textwidth]{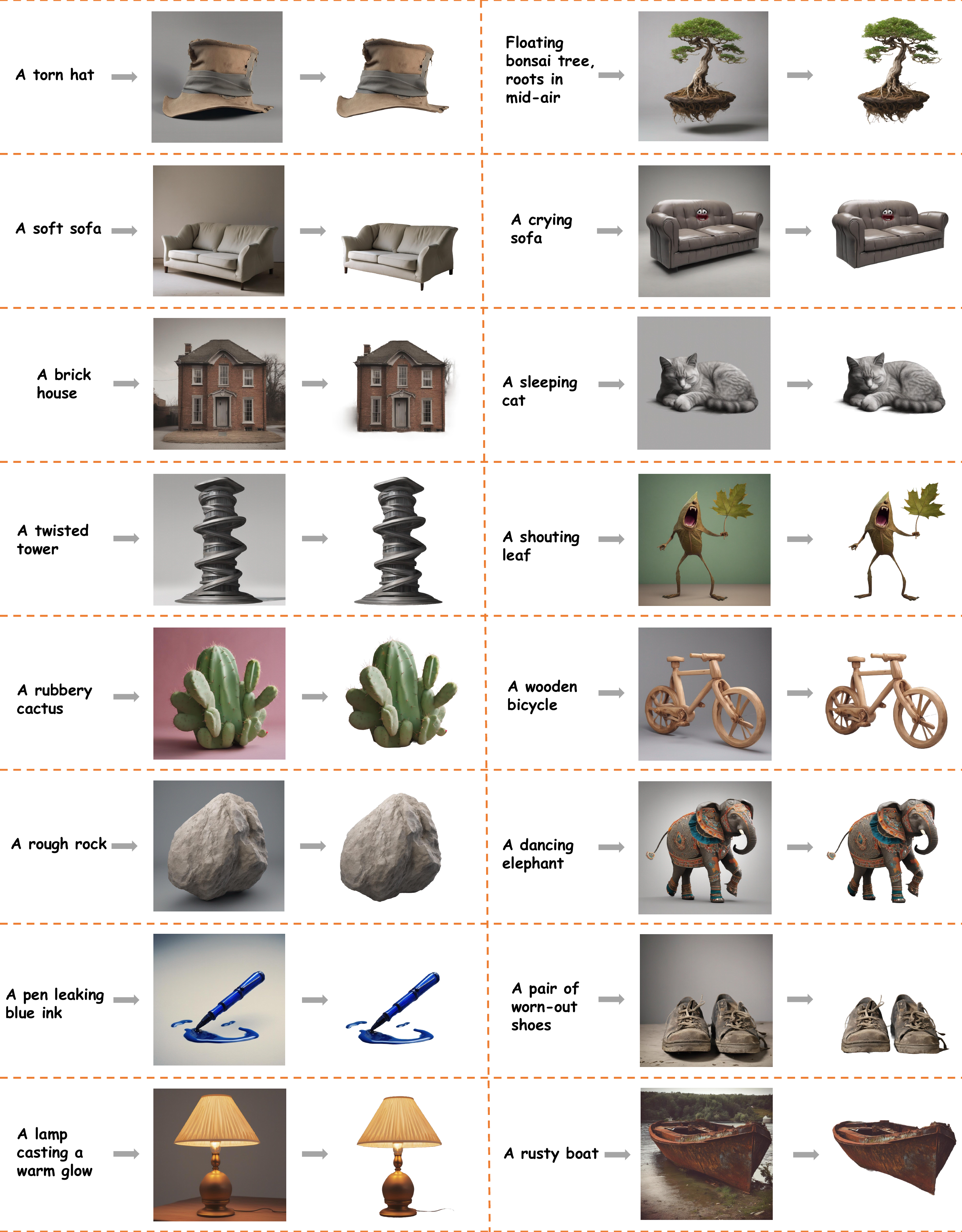}
  \vspace{-1mm}
  \caption{Testset: 100 3D Assets from Stable Diffusion (2).}
  \label{fig_100sd_2}
  \vspace{30mm}
\end{figure} 

\section{Additional Experiments}
\label{sec:D_Additional_Experiments}

In this section, we supplement our experiments. In \cref{sec:D_1_Compare with More Multi-view Reconstruction}, we compare our method's effectiveness with more recent multi-view reconstruction techniques. In \cref{sec:D_2_More Examples}, similar to \Cref{fig_exp_fill_and_style}, We present additional examples of Tailor3D, showcasing our ability to customize and edit objects.

\begin{figure}[!t]
  \centering
  \includegraphics[width=\textwidth]{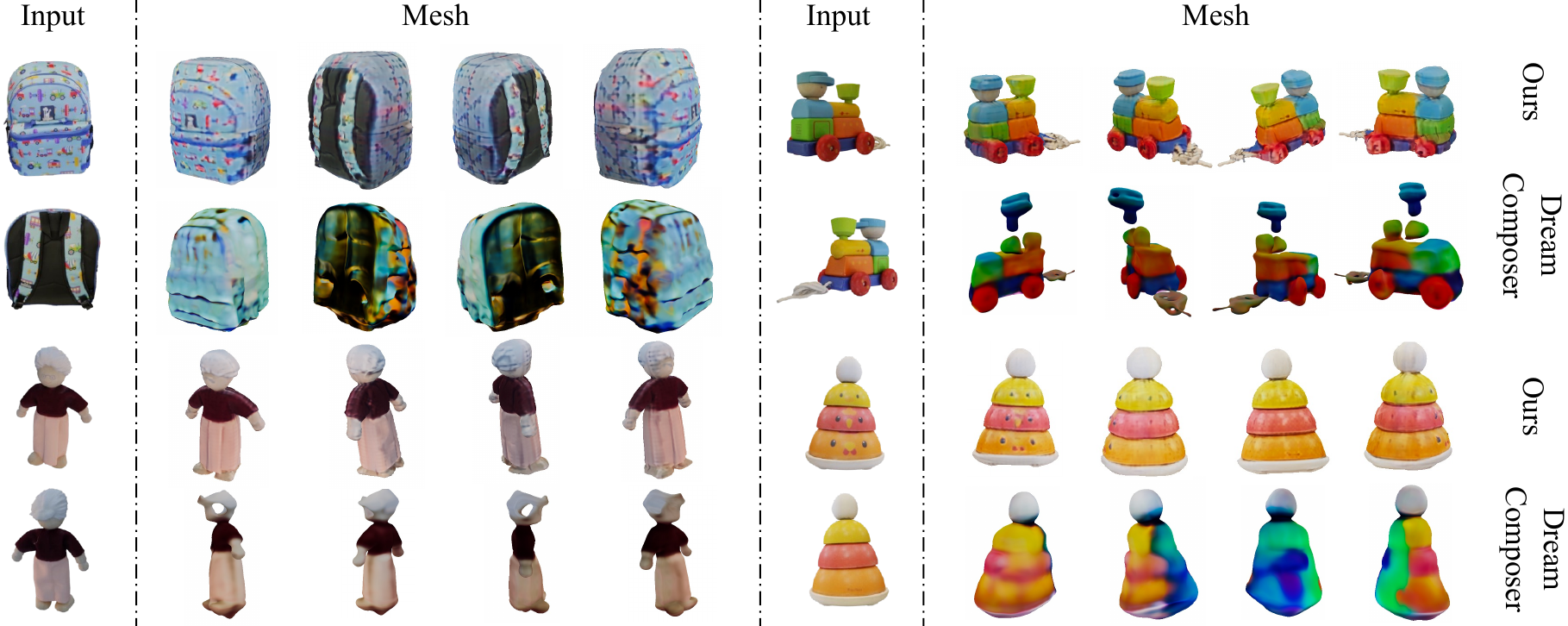}
  % \vspace{-10pt}
  \caption{\textbf{Compare with Dreamcomposer.} Here, we present a comparison with the multi-view DreamComposer~\cite{DreamComposer}. In this comparison, we provide Tailor3D with ground-truth RGB images for the back side. It can be observed that Tailor3D exhibits more detailed texture features and avoids defects such as holes.}
  \label{fig_dreamcomposer}
  % \vspace{-10pt}
\end{figure}

\begin{figure}[t]
  \centering
  \includegraphics[width=\textwidth]{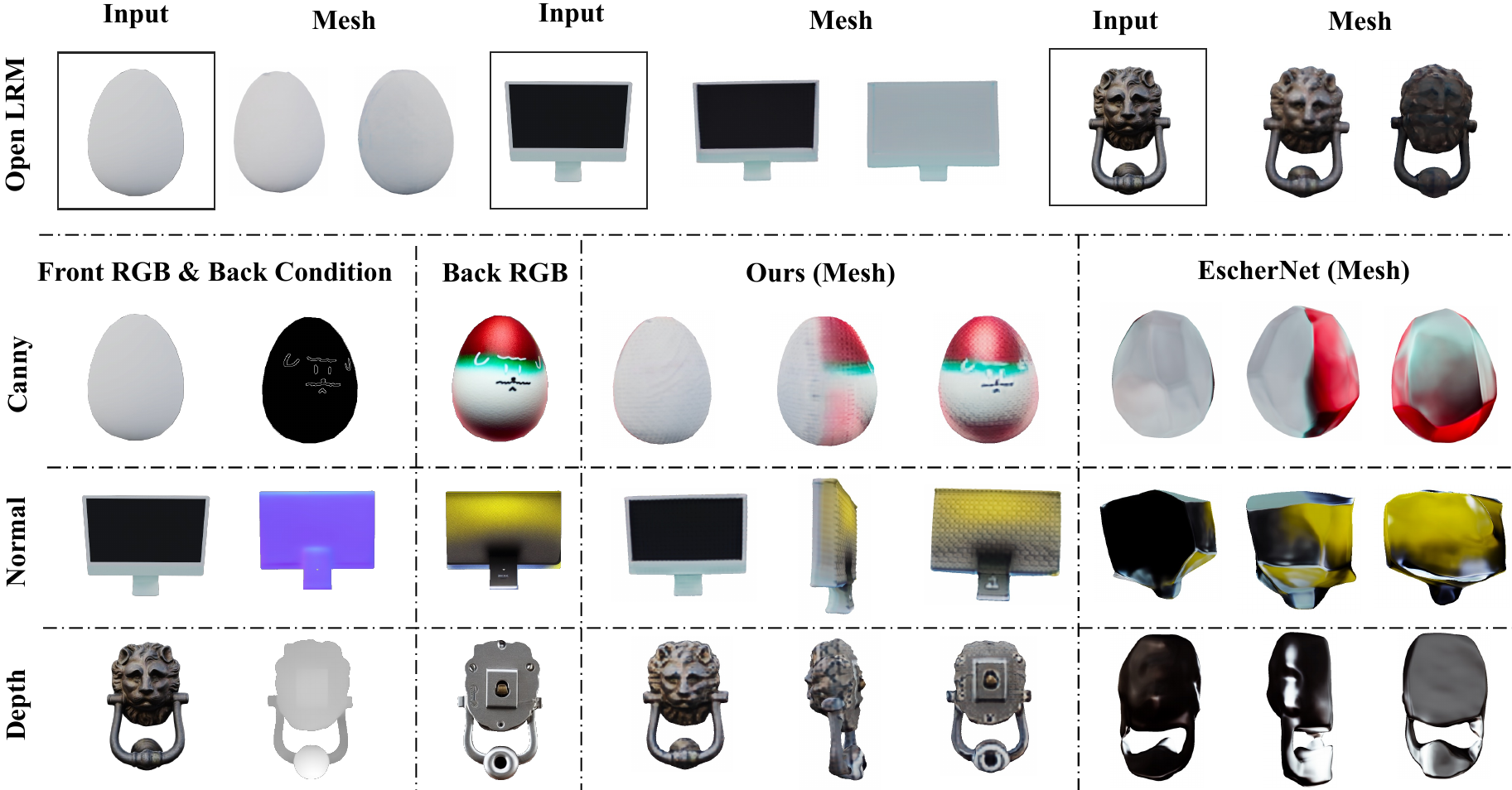}
  % \vspace{-10pt}
  \caption{\textbf{Compare withh multi-view input model EscherNet~\cite{EscherNet}. } Our created mesh excels beyond other methods, delivering superior speed and quality.}
  \label{fig_eschernet}
\end{figure} 

\subsection{Comparison with More Multi-view Reconstructions}
\label{sec:D_1_Compare with More Multi-view Reconstruction}
In the main paper, we compared earlier 3D generation methods like Wonder3D~\cite{Wonder3d}, TriplaneGaussian~\cite{TriplaneGaussian}, and LGM~\cite{LGM}, most of which were focused on image-to-3D generation. Conversely, approaches like Dreamcomposer~\cite{DreamComposer} and EscherNet~\cite{EscherNet} aimed to complement additional viewpoints. It's worth noting here the test set is from GSO30~\cite{GSO30} and Objaverse~\cite{Objaverse} datasets instead of the 100 SD test set used in the main paper. Dreamcomposer and EscherNet are optimization-based methods, thus requiring several minutes to generate 3D results. In contrast, Tailor3D only needs 5 seconds to produce superior 3D reconstruction results.

\textbf{Comparison with Dreamcomposer.}
DreamComposer is built on SyncDreamer~\cite{SyncDreamer}, allowing it to accept inputs from multiple viewpoints and fill in missing information for all sides except the back. In our experimental results (see \cref{fig_dreamcomposer}), we adjusted the back input to be the RGB image of the ground-truth back side for comparison purposes. That is, we provided Tailor3D and Dreamcomposer with pictures of the front and back of the object, which could have been more perfectly consistent. We found that Tailor can generate superior mesh results compared to DreamComposer. DreamComposer tends to exhibit more defects in its reconstructions.

\textbf{Comparison with EscherNet.}
EscherNet is a multi-view conditional diffusion model for viewpoint synthesis. It learns implicit and generative 3D representations combined with Camera Position Encoding (CaPE). EscherNet can generate more consistent images and has higher reconstruction quality. In this experiment, we provided EscherNet with 16 viewpoints, while our Tailor3D had only the front and back viewpoints. Even in this scenario, our approach still has a significant advantage and obtains better mesh results. This further demonstrates that our method using only two views for reconstruction can achieve better results.

\subsection{More Examples}
\label{sec:D_2_More Examples}
Here, we showcase more qualitative examples, including 3D style transfer, style fusion, and 3D generative fill. We demonstrate the model's ability to transform overall styles as well as perform localized editing. These examples are visually stunning, showcasing the potential for industrial applications.

\begin{figure}[t!]
  \centering
  \includegraphics[width=0.99\textwidth]{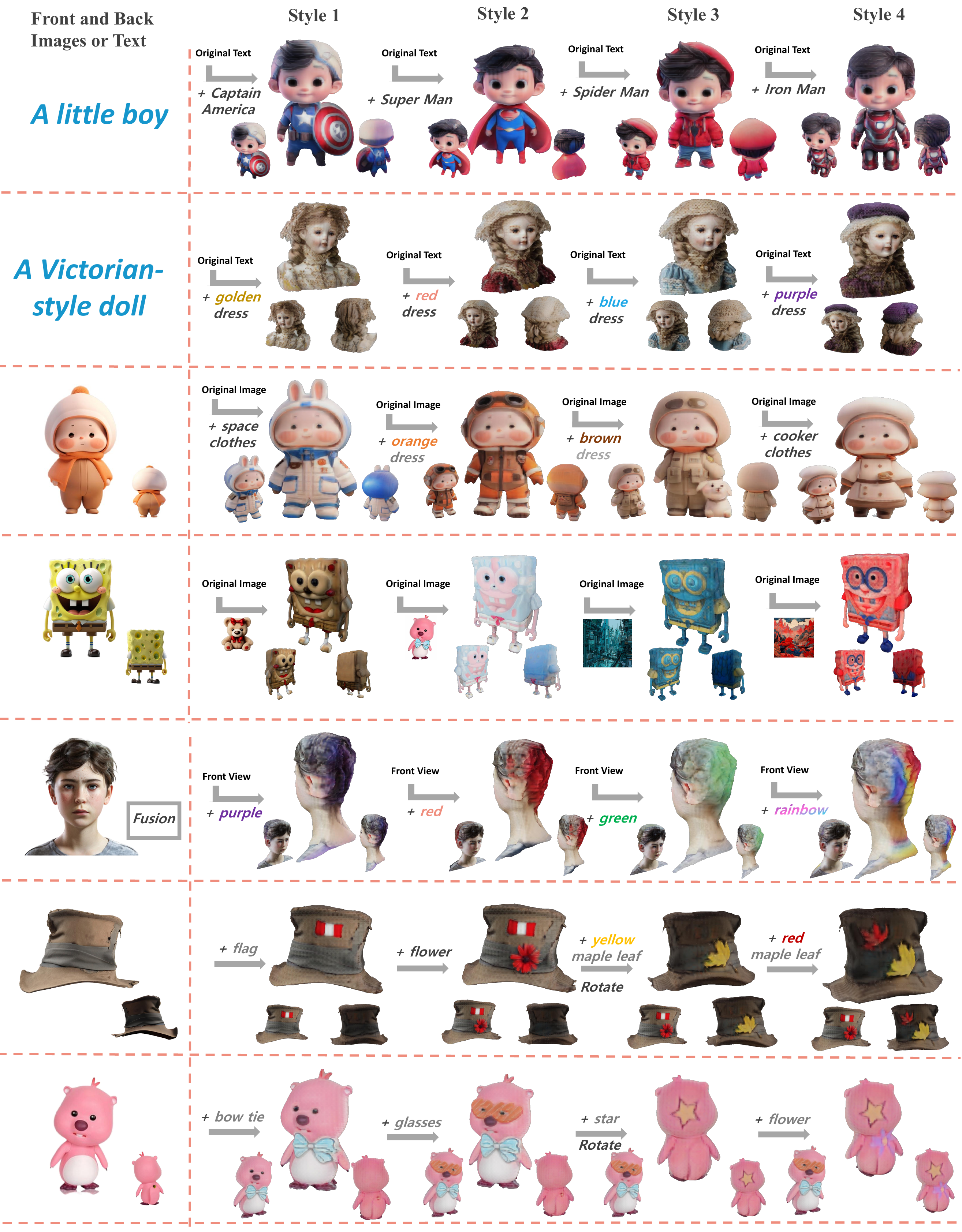}
  \vspace{-1mm}
  \caption{More Examples about Tailor3D.}
  \label{fig_exp_supdemo}
  \vspace{3mm}
\end{figure} 

\section{Broad Impacts}
\label{sec:E_Broader_Impacts}
In this section, we emphasize Tailor3D's broad societal impact. Our method is practical, as demonstrated by our qualitative experimental results.

\textbf{Academic Impact}: Tailor3D introduces a novel approach to 3D generation, starting from a single image or text and using multi-view diffusion to generate Canonical views. By delegating editing to 2D images and reconstruction to 3D, our method provides a direction for future exploration in 3D generation within the academic community.

\textbf{Industrial Impact}: Tailor3D is a practical paper, as evidenced by our qualitative experimental results. Our fine-grained operations result in highly editable and applicable outcomes. Our motivation is rooted in considering user input and requirements. Furthermore, each step of our method operates at a sub-second level, making it applicable to various industrial scenarios.

\textbf{Social Impact}: Tailor3D can be applied in animation production, 3D game development, and other fields, significantly reducing production costs and sparking a wave of creative content creation. This democratization of creation allows society to enjoy the fruits of AI development better.

\end{document}